\documentclass[journal,12pt, draftclsnofoot,onecolumn]{IEEEtran}
\IEEEoverridecommandlockouts
\usepackage{setspace}
\usepackage{cite}
\usepackage{amsmath,amssymb,amsfonts,mathtools,bm}
\usepackage{amsthm}
\usepackage{algorithm}
\usepackage{graphicx}
\usepackage{textcomp}
\usepackage{xcolor,float}
\usepackage{tabularx}
\usepackage{textcomp}
\usepackage{diagbox}
\usepackage{svg}
\usepackage{booktabs}
\usepackage{subfigure}
\usepackage{hyperref}

\newtheorem{problem}{Problem}
\usepackage{algpseudocode}
\usepackage{graphicx}
\usepackage{makecell}
\usepackage{amsthm}
\usepackage{caption}
\usepackage{sidecap}
\usepackage{microtype}
\usepackage{booktabs} 
\usepackage{multirow}
\usepackage{float}
\usepackage{adjustbox} 
\usepackage{amsmath}
\usepackage{amssymb}
\usepackage{colortbl}
\usepackage{makecell}
\usepackage{float}
\usepackage{url}
\usepackage{balance}
\usepackage{bbding}
\usepackage{hyperref}
\usepackage{graphicx}
\usepackage{sidecap}
\usepackage{microtype}
\usepackage{graphicx}
\usepackage{subfigure}
\usepackage{booktabs} 
\usepackage{multirow}
\usepackage{array}
\usepackage{subcaption}
\usepackage{graphicx}
\usepackage{makecell}
\usepackage{amsmath}
\usepackage{amssymb}

\usepackage{tikz}
\usepackage{changes}
\usepackage{ifthen}
\newboolean{showchanges}
\setboolean{showchanges}{false}  % True for blue

\newcommand{\add}[1]{%
    \ifthenelse{\boolean{showchanges}}%
        {\textcolor{blue}{#1}}
        {#1\relax}
}

\definecolor{lime}{HTML}{A6CE39}
\DeclareRobustCommand{\orcidicon}{%
    \begin{tikzpicture}
    \draw[lime, fill=lime] (0,0) 
    circle [radius=0.16] 
    node[white] {{\fontfamily{qag}\selectfont \tiny ID}};    \draw[white, fill=white] (-0.0625,0.095) 
    circle [radius=0.007];    \end{tikzpicture}
    \hspace{-2mm}}
\foreach \x in {A, ..., Z}{%
    \expandafter\xdef\csname orcid\x\endcsname{\noexpand\href{https://orcid.org/\csname orcidauthor\x\endcsname}{\noexpand\orcidicon}}
    }
\allowdisplaybreaks
\renewcommand{\baselinestretch}{1}

\begin{document}
\doublespacing

\title{RadioDiff: An Effective Generative Diffusion Model for Sampling-Free Dynamic Radio Map Construction}

\author{
Xiucheng Wang\orcidA{},~\IEEEmembership{Student Member,~IEEE,}
Keda Tao\orcidB{},
Nan Cheng\orcidC{},~\IEEEmembership{Senior Member,~IEEE,}
Zhisheng Yin\orcidD{},~\IEEEmembership{Member,~IEEE,}
Zan Li\orcidE{},~\IEEEmembership{Senior Member,~IEEE,}
Yuan Zhang\orcidF{},~\IEEEmembership{Member,~IEEE,}
Xuemin (Sherman) Shen\orcidG{},~\IEEEmembership{Fellow,~IEEE}

\thanks{ }% <-this % stops a space
\thanks{
\par Xiucheng Wang, Keda Tao, Nan Cheng, Zhisheng Yin, and Zan Li are with the State Key Laboratory of ISN and School of Telecommunications Engineering, Xidian University, Xi’an 710071, China (e-mail: xcwang\_1@stu.xidian.edu.cn; KD.TAO.CC@outlook.com; dr.nan.cheng@ieee.org; \{zsyin, zanli\}@xidian.edu.cn). Xiucheng Wang and Keda Tao contribute equally. \textit{Nan Cheng is the corresponding author}.
\par Yuan Zhang is with the School of Computer Science and Engineering, University of Electronic Science and Technology of China, Chengdu 610054, China (e-mail: zy\_loye@126.com).
\par Xuemin (Sherman) Shen is with the Department of Electrical and Computer Engineering, University of Waterloo, Waterloo, N2L 3G1, Canada (e-mail: sshen@uwaterloo.ca).
}

}

    \maketitle

\IEEEdisplaynontitleabstractindextext

\IEEEpeerreviewmaketitle

\begin{abstract}
Radio map (RM) is a promising technology that can obtain pathloss based on only location, which is significant for 6G network applications to reduce the communication costs for pathloss estimation. However, the construction of RM in traditional is either computationally intensive or depends on costly sampling-based pathloss measurements. Although the neural network (NN)-based method can efficiently construct the RM without sampling, its performance is still suboptimal. This is primarily due to the misalignment between the generative characteristics of the RM construction problem and the discrimination modeling exploited by existing NN-based methods. Thus, to enhance RM construction performance, in this paper, the sampling-free RM construction is modeled as a conditional generative problem, where a denoised diffusion-based method, named RadioDiff, is proposed to achieve high-quality RM construction. In addition, to enhance the diffusion model's capability of extracting features from dynamic environments, an attention U-Net with an adaptive fast Fourier transform module is employed as the backbone network to improve the dynamic environmental features extracting capability. Meanwhile, the decoupled diffusion model is utilized to further enhance the construction performance of RMs. Moreover, a comprehensive theoretical analysis of why the RM construction is a generative problem is provided for the first time, from both perspectives of data features and NN training methods. Experimental results show that the proposed RadioDiff achieves state-of-the-art performance in all three metrics of accuracy, structural similarity, and peak signal-to-noise ratio. The code is available at \url{https://github.com/UNIC-Lab/RadioDiff}.
\end{abstract}

\begin{IEEEkeywords}
radio map, denoise diffusion model, generative problem, wireless network.

\end{IEEEkeywords}

\section{Introduction}
In wireless networks, pathloss quantifies the attenuation of signal strength between a pair of sender and receiver caused by free-space propagation loss and interactions of radio waves with obstacles \cite{liu2023exploiting, jiwei2024channel,10648594}, which is critical to be measured for wireless resource management \cite{cheng2019space,wang2022joint,wang2023scalable}. Traditionally, the pathloss measurement usually depends on pilot transmission and signal processing \cite{wang2020indoor}. However, the dramatic increase of network nodes and antennas has led to the challenge of estimating high-dimensional channels \cite{dang2020should}, resulting in significant costs in training and feedback overhead, as well as signal processing complexity \cite{shen2023toward}. This issue further deteriorates in high-mobility scenarios with short channel coherence times for low-latency applications \cite{6g,liu2020uplink}. Meanwhile, the upcoming 6G networks will introduce a large variety of node types, including passive equipment such as intelligent reflective surfaces (IRS) \cite{you2020wireless}, which cannot actively transmit pilots or engage in digital signal processing to measure the pathloss. 
% Additionally, as an important component of 6G networks, unmanned aerial vehicle (UAV) networks often require planning of UAV flight trajectories to optimize network performance, and it is highly helpful to have advanced knowledge of user path loss in upcoming areas to facilitate the reasonable planning of flight trajectories, thereby enhancing the performance of UAV networks \cite{wang2022joint,cheng2019space}.

The emergence of these new scenarios makes it necessary to efficiently obtain pathloss using easy-to-be-obtained information without pilot transmission and signal processing. Consequently, radio map (RM) technology has been developed, by which the pathloss can be acquired just through location information \cite{zeng2024tutorial}. Traditional RM construction methods can be categorized into two types: (1) sampling-based approaches that rely on sampling position measurements (SPM) within the RM region, which are then used for interpolation or solving specific least squares problems to construct the RM \cite{cover1967nearest,breidt2000local,qiu2024channel}, and (2) sampling-free methods which are achieved through environmental 3D modeling and electromagnetic ray tracing (ERT) \cite{oh2004mimo}. However, both methods face their own inherent challenges. The sampling-based methods require SPM of the RM construction area, with too few or inaccurate measurements leading to poor construction quality, while a large number of high-precision measurements significantly increase RM construction costs. Moreover, such methods cannot be used to construct RM in never-to-reach regions, limiting its applicability, such as UAV trajectory plan. On the other hand, the ERT-based method, while avoiding measurement costs, is burdened with high computational complexity and struggles to achieve RM construction within acceptable timeframes. Furthermore, both methods are restricted to the construction of static RM (SRM) due to their construction principles, which do not account for the real-time transformation of elements affecting pathloss. Consequently, changes in pathloss resulting from factors such as vehicle motion or alterations in reflection diameters are not reflected in the RM. The sampling-based method requires multiple sampling points to be measured across a wide area, making it impractical to arrange sufficient measurement equipment for simultaneous pathloss collection. Additionally, the time-series measurement of different sampling points/sets leads to data collection at different times, rendering real-time RM construction unattainable \cite{cover1967nearest,breidt2000local}. Similarly, the ERT-based method, which fundamentally relies on ray tracing of a static 3D scene, typically involves calculations that take several minutes or even longer, rendering it unsuitable for RM construction with dynamic environmental features \cite{oh2004mimo}.

\begin{figure}[t]
\captionsetup{font={small}, skip=16pt}
    \centering
    \vspace{-9pt}
    \subfigure[The illustration of SRM.]
    {
       \centering
       \includegraphics[width=0.405\columnwidth]{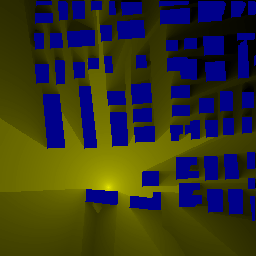}\label{srm}
    }
    \subfigure[The illustration of DRM.]
    {
       \centering
       \includegraphics[height=0.405\columnwidth]{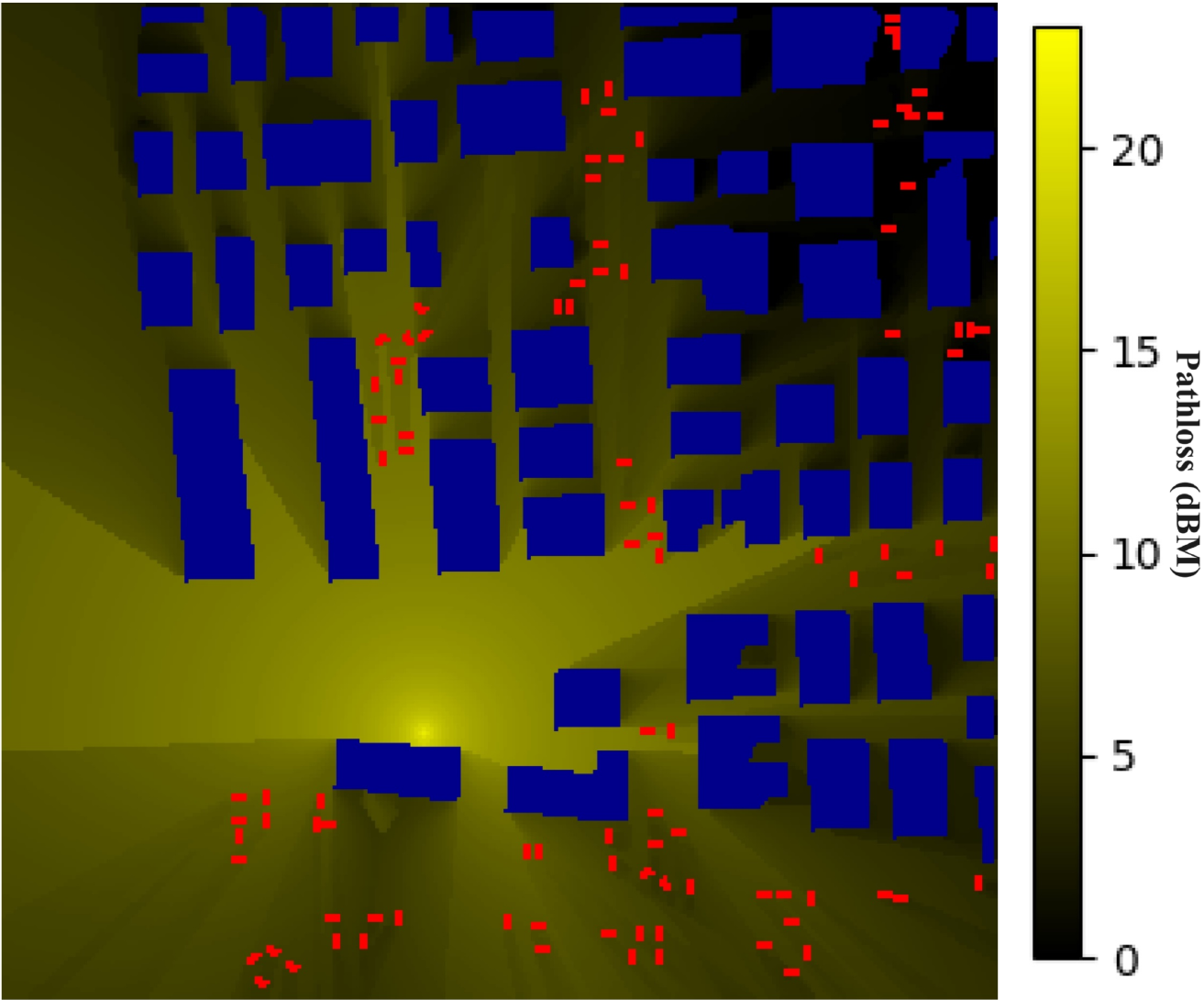}\label{drm}
    }
    \caption{Illustration of the RM, where the {\color[HTML]{f08a5d} \textbf{yellow}} elements represent the heatmap of pathloss; the brighter the yellow, the higher the pathloss. The \textcolor{red}{\textbf{red}} elements denote vehicles, and the \textcolor{blue}{\textbf{blue}} elements are static buildings. Since, static buildings can completely block electromagnetic signals from entering their interiors, resulting in an internal pathloss of zero, to intuitively represent the impact of static buildings on the RM, we have colored them blue.}
    \label{fig-example}
    \vspace{-9pt}
\end{figure}

To address these challenges, numerous researchers have exploited neural networks (NNs) with rapid inference capabilities to facilitate RM construction. The most notable pioneering work is the RadioUNet that leverages the U-Net, which is a classical architecture for image-to-image tasks, for RM construction \cite{levie2021radiounet}. Although some NN-based methods have demonstrated better performance than traditional SPM-based methods in RM construction, their performance, especially in details construction for the dynamic RM (DRM) with multiple dynamic obstacles\footnote{It is important to note that in this paper, we construct the RM based on a snapshot of the environment, without considering the Doppler effect caused by vehicle motion on the pathloss. Consequently, the main difference between DRM and SRM lies in the snapshot time interval and whether the influence of small-scale obstacles on the propagation of electromagnetic rays is considered.}, is poor. Therefore, the majority of NN-based methods focus on RM construction in static environments, neglecting the exploration of RM construction in dynamic environments \cite{levie2021radiounet,li2022radionet,chen2023graph,zhang2023rme}. This is primarily attributed to two challenges: the heterogeneous propagation characteristics of electromagnetic waves and the complexity of RM texture features. The RM construction in static environments typically only concentrates on the influence of buildings on electromagnetic ray propagation, since it can be assumed that all objects affect electromagnetic rays in the same manner. In the context of DRM construction, the impact of moving objects, such as vehicles, on electromagnetic rays must be taken into account. As shown in Fig.~\ref{fig-example}, the influence of vehicles on electromagnetic ray propagation markedly differs from that of buildings. Different from static obstacles with large shapes and high heights, which can completely block direct electromagnetic rays from the base station (BS) to their surfaces, dynamic obstacles such as vehicles, due to their low height and small size, cannot entirely obstruct the electromagnetic signal. This partial blockage results in a reduction of the pathloss rather than a complete obstruction, thereby increasing the complexity of RM construction and the texture features than SRM. Moreover, existing NN-based methods mainly train the NN through supervised learning in a discriminative style, whose objective is to minimize the mean square error (MSE) between the predicted RM and the ground truth. However, it has been demonstrated in \cite{austin2021structured, wang2004image} that while the widely used MSE metric in discriminative training methods enhances the convergence speed of NN training, it degrades the NN's ability to capture the elaborately detailed features of the data, resulting in blurred edges. \added{As shown in Fig.~\ref{fig-example}, RM (especially DRM) typically contains more detailed texture features. Neural networks trained with MSE-based supervised learning methods struggle to extract these detailed features.}

The fundamental reason for the above challenges is the misalignment between the generative problem attributes of the RM construction and the discriminant methodology. RM construction shows the characteristics of generative problems in both data features and NN training methods, while the existing sample-free NN-based methods use discriminant methods to construct RM, which inevitably limits the construction performance of RM. Specifically, in terms of data features, neither the value nor location of the elements of the pathloss to be predicted exist in the input environmental data, thus NN is required to generate pathloss from the raw data. Moreover, since the elements in the RM are not discrete values, it is almost impossible to construct the RM by dividing a finite number of hyperplanes to classify the elements in the environmental data as a discriminative problem \cite{vapnik1998statistical, wang2024reliable}. Therefore, the use of discriminant training methods to generate multiple hyperplanes in the latent space using NN to predict pathloss will inevitably lead to poor performance. Furthermore, the self-supervised training method that uses data which is partially masked as input to train NN to predict masked information is mainly used in the training of generative models \cite{goodfellow2016deep}. In the context of RM construction, the data representing environmental information, can be regarded as the RM whose pathloss is masked, and the NN-based RM construction methods use these masked RM data to recover the pathloss, which is a self-supervision training method. Based on the above analysis, RM construction is a generative problem, so from the perspective of the alignment of the problem and the methodology, a generative method should be used to effectively construct the RM. Moreover, the generative training methods enable NN training both using generative loss and MSE loss, where the NN's capacity to extract intricate texture features of RM can be enhanced. 
\add{
Generative adversarial networks (GANs) have been extensively studied in RM reconstruction; however, their potential is often hindered by extreme instability during training. In contrast, the diffusion model offers a novel solution to the RM reconstruction problem. Geographic maps and RM contain abundant high-frequency sharp edges. Directly feeding these into NN-based models can result in erratic predictions, noise, and artifacts. Diffusion models, however, demonstrate superior capability in capturing this edge information and predicting conditions.
}Therefore, to achieve high-performance RM construction by taking advantage of the generative NN training method, in this paper, the RM construction problem is modeled as a conditional generative problem, where a generative diffusion-based method, named RadioDiff, is proposed to construct the RM effectively \footnote{In this paper, the terms ``diffusion model" and ``denoise diffusion model" are used interchangeably, both referring to methods for generating the required data from noise input. The diffusion process involves diffusing the data into noise, while the denoising process reconstructs the data from this noisy input.}. The main contributions of this paper are as follows.
\begin{enumerate}
    \item For the first time, the sampling-free RM construction problem is modeled as a conditional generative problem, where the location of the BS, and the environment features are used as prompts for conditional generation. In addition, theoretically, a detailed analysis of why the RM construction is a generative problem from the perspective of data features and NN training methods is provided for the first time.
    \item To the best of our knowledge, the diffusion-based generative model is first used for effective RM construction. Moreover, to enhance the performance and the inferencing efficiency of the diffusion model, a decoupled diffusion model is used in this paper.
    \item To enhance the diffusion model's capability to extract dynamic environmental features, the prompts of the static and dynamic environmental features are represented by two matrices, respectively. Additionally, the adaptive fast Fourier transform module is employed to enhance the diffusion model's ability to extract high-frequency information resulting from dynamic environmental features in the data.
    \item The experimental results demonstrate that the proposed RadioDiff achieves state-of-the-art (SOTA) RM construction performance in all three metrics of accuracy, structural similarity, and peak signal-to-noise ratio.
\end{enumerate}
The remainder of this paper is organized as follows. We first overview the related works of RM construction and give preliminaries of the diffusion model in section-\ref{sec-2}, then the RM construction problem is formulated and analyzed in section-\ref{sec-3}. In section-\ref{sec-4} the details of the proposed RadioDiff are introduced, while in section-\ref{sec-5} the experimental results are given. The section-\ref{sec-7} concludes our work. The notations are shown in Table \ref{tab-notation}.

\begin{table}
    \centering
    \caption{Notation Table}
    \resizebox{0.5\linewidth}{!}{
    \begin{tabular}{c|c}
    \hline
       Variables  & Definition\\\hline
        $\bm{x}_{t}$ & Noise data after $t$ times of diffusion. \\ 
       $\bm{z}_{t}$  & The feature map of $\bm{x}_{t}$. \\
       $\bm{\hat{z}}_{t}$  & The feature map predicted by NN. \\
       $\bm{\mu}_{\bm{\theta}}(\cdot)$  & The denoise NN with parameters $\bm{\theta}$.\\
       $\bm{w}$ & The parameters of AFT.\\
       $\mathcal{E}$ & VAE encoder.\\
       $\mathcal{D}$ & VAE decoder.\\
       \added{$q(\cdot)$} & \added{The probability of forward diffusion process.}\\
       \added{$p(\cdot)$} & \added{The probability of inverse denoising process.}\\
       $\bm{\phi}(\cdot)$ & Flattened operator.\\
       $\bm{\nu}(\cdot)$ & Trainable projection function.\\
       $\mathcal{F}(\cdot)$ & The FFT operation.\\
       $\mathcal{F}^{-1}(\cdot)$ & The inverse FFT operation.\\
       $\bm{P}$ & The ground truth of RM.\\
       $\bm{\hat{P}}$ & The RM predicted by NN.\\
       $\bm{H}_{s}$ & The static obstacle distribution matrix.\\
       $\bm{H}_{d}$ & The dynamic obstacle distribution matrix.\\
       $\bm{R}$ & The location of BS.\\
       $\bm{C}$ & The prompt of the diffusion model.\\
       $\bm{I}$ & The identity matrix.\\
       $\mathcal{N}(\cdot,\cdot)$ & The Gaussian stochastic distribution.\\
       $\mathbb{R}^{N}$ & $N$-dimensional real space.\\
       $\bm{\epsilon}$ & Gaussian stochastic variable as added noise.\\
       $T$ & Maximum number of adding noises.\\
       $\alpha,\beta,\gamma,\delta$ & Hyper-parameters.\\

       \hline
       \end{tabular}
    }
    \label{tab-notation}
\end{table}

\section{Preliminaries and Related Work}\label{sec-2}
\subsection{Radio Map Construction}
The construction of RM can be categorized into two primary types: sampling-based and sampling-free. The sampling-based methods primarily utilize SPM to obtain pathloss in specific areas for interpolation. Although these methods are independent of knowing environmental details and BS location, they require pathloss measurement in the targeted regions for RM construction. Among these methods, the K-nearest neighbors technique obtains RM data for other locations by weighted averaging the pathloss values of the K nearest sparse measurement points \cite{cover1967nearest}. Additionally, local multinomial regression is commonly employed in sampling-based RM construction. This approach determines the pathloss at a current point by solving a least squares problem that involves the pathloss values of nearby sparse measurement points \cite{breidt2000local}. To further enhance the quality of sampling-based RM construction, Kriging interpolation has been proposed. This method treats RM construction as a stochastic process modeling and prediction problem based on a covariance function, thereby improving the accuracy of RM construction \cite{qiu2024channel}.

The aforementioned sampling-based methods face two significant challenges: dependence on SPM and low construction accuracy \cite{levie2021radiounet,li2022radionet}. Consequently, sampling-free RM construction methods, based on NN, have attracted the attention of researchers. These methods typically require knowledge of the environmental features, such as obstacle locations, heights, and the positions of BS, to construct RMs. A representative example is the RadioUNet, which is derived from the classic U-Net framework used in image-to-image tasks \cite{levie2021radiounet}. This approach leverages the U-Net framework, training NN using the MSE between the generated RM and the ground truth, yielding impressive results \cite{levie2021radiounet}. Inspired by RadioUNet and the success of the Vision Transformer (ViT) led to the development of Radionet, based on the transformer architecture for RM construction \cite{li2022radionet}. Additionally, more complex NN architectures, such as graph neural networks (GNN) \cite{chen2023graph}, with stronger feature extraction capabilities, have been employed in RM construction. However, these approaches generally treat sampling-free RM construction as a discriminative supervised learning task. Although RME-GAN attempts to introduce generative adversarial networks (GAN) methods into RM construction, it is not sampling-free, as it relies on SPM within the construction area \cite{zhang2023rme}. Different from existing NN-based RM construction methods, in this paper, the sampling-free RM construction is modeled as a conditional generative problem, and a diffusion-based method is proposed which significantly improves the RM construction performance.

\subsection{Diffusion Model}
The diffusion model is a category of generative models based on Markov chains, that progressively restore data through a learned denoising process. This model has emerged as a strong competitor to the GAN in various generative tasks, such as computer vision \cite{lin2023diffbir}, and natural language processing \cite{austin2021structured}. Moreover, diffusion models exhibit significant potential in perception tasks, including image segmentation, object detection, and model-based reinforcement learning (RL) \cite{LDM}. For example, in \cite{10638123}, the authors effectively utilized diffusion model-based soft actor-critic algorithms for optimal contract design. Besides, the authors in \cite{10628023} innovatively leveraged diffusion model-based deep deterministic policy gradient algorithms for optimal Stackelberg game solutions. In the diffusion model, there are two procedures which are the forward diffusion procedure where the raw data is diffused into noise, and the reversed denoise procedure where an NN is used to remove the noise adding to the raw data, thus generating raw data from noise.
\begin{figure*}[t]
\centering
\includegraphics[width=0.95\textwidth]{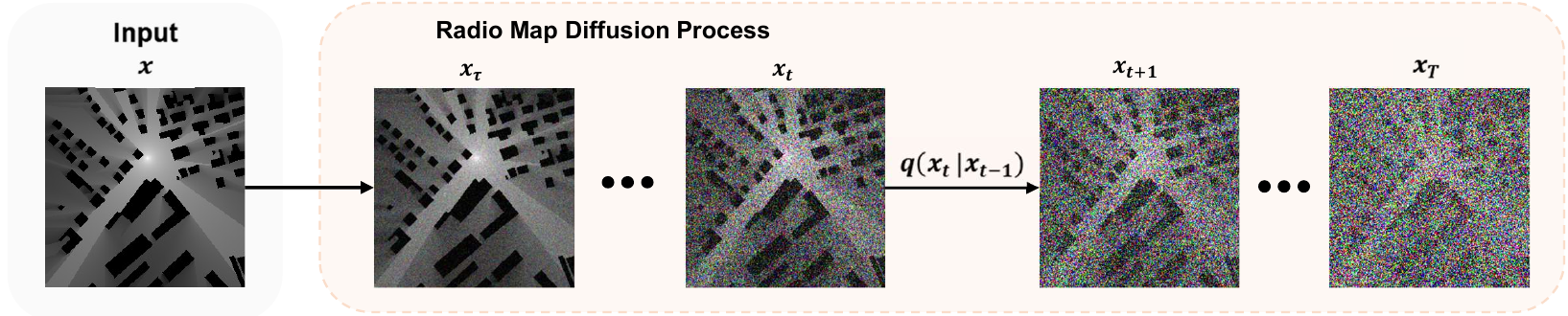} % Reduce the figure size so that it is slightly narrower than the column.
\caption{ The diffusion procedure of RM, where in diffusion procedure the RM is diffused into noise, and in the denoising procedure the RM is revealed from the noise and prompt.}
\label{fig_diffusion}
\end{figure*}

\subsubsection{Forward Diffusion Procedure}
From a probabilistic modeling standpoint, the essence of generative models lies in training them to produce data $\hat{\bm{x}}\sim p_\theta(\hat{\bm{x}})$ that mirror the distribution of the training data $\bm{x}\sim p_{t}(\bm{x})$. The denoising diffusion probability model (DDPM) employs two Markov chains: a forward chain that converts data into noise, and a backward chain that reverts the noise to data. In a formal context, given the data $\bm{x}_0$, the progression of a forward Markov chain is realized by generating a series of stochastic variables $ \bm{x}_1, \bm{x}_2, \ldots, \bm{x}_T $, which evolve following the transition kernel $ q(\bm{x}_t | \bm{x}_{t-1}) $. By employing the chain rule of probability in conjunction with the Markovian property, it is possible to deconstruct the joint probability distribution of $ \bm{x}_1, \bm{x}_2, \ldots, \bm{x}_T $ given $ \bm{x}_0 $, expressed as $ q(\bm{x}_1, \ldots, \bm{x}_T | \bm{x}_0) $, into an appropriate factorial form.
\begin{equation}q(\bm{x}_1,\ldots,\bm{x}_T\mid\bm{x}_0)=\prod_{t=1}^Tq(\bm{x}_t\mid\bm{x}_{t-1}).\end{equation}
Thus, a forward noising process which produces latent $\bm{x}_t$ through $\bm{x}_0$ by adding Gaussian noise at time $t\in\{0,1,\cdots,T\}$ can be defined as follows.
\begin{equation}
q\left(\bm{x}_t\mid \bm{x}_{t-1}\right)=\mathcal{N}\left(\sqrt{1-\beta_t} \bm{x}_{t-1},\beta_t\bm{I}\right),
\end{equation}
where $T$ and $\beta_t\in(0,1)$ are the total number of diffusion iterations and hyper-parameter of the variance scaling factor, respectively. By setting $ \alpha_t = 1 - \beta_t $ and $ \bar{\alpha}_t = \prod_{s=0}^{t} \alpha_s $, the distribution of $\bm{x}_{t}$ condition on the $\bm{x}_0$ can be obtained as follow.
\begin{equation}
q(\bm{x}_t\mid\bm{x}_0)=\mathcal{N}(\sqrt{\bar{\alpha}_t}\bm{x}_0,(1-\bar{\alpha}_t)\bm{I}),\label{diffused}
\end{equation}
\begin{equation}
\bm{x}_t=\sqrt{\bar{\alpha}_t}\bm{x}_0+\sqrt{1-\bar{\alpha}_t}\bm{\epsilon},
\end{equation}
where $\mathcal{N}(\cdot,\cdot)$ is Gaussian distribution, $\bm{I}$ is identity matrix, and $\bm{\epsilon}\sim\mathcal{N}(0,\bm{I})$. 
\add{In addition, Eq. (4) describes how the noisy data $\bm{x}_t$ is generated by combining the original input $\bm{x}_0$ with a Gaussian noise term $\bm{\epsilon}$. The term $\sqrt{\bar{\alpha}_t} \bm{x}_0$ represents the scaled contribution of the original data, while $\sqrt{1 - \bar{\alpha}_t} \bm{\epsilon}$ represents the amount of noise added at each step. As $t$ increases, the influence of the noise term grows, resulting in progressively noisier data.}

\subsubsection{Reversed Denoise Procedure}
For the data generation, DDPM initially creates unstructured noise vectors from the prior distribution, subsequently removing the noise through a learnable Markov chain operated in reverse temporal order. To achieve this, the reverse process can be formulated as follows.
\begin{equation}
    p_{\bm{\theta}}(\bm{x}_{t-1}|\bm{x}_t)=\mathcal{N}(\bm{\mu}_{\bm{\theta}}(\bm{x}_t,t),\beta_t\bm{I})\label{reverse-2}
\end{equation}
where $\beta_{t}$ is a hyper-parameter, $\bm{\theta}$ is the parameter of NN $\bm{\mu}_{\bm{\theta}}$. Applying a trained NN $\bm{\mu}_{\bm{\theta}}$, we can iteratively denoise $\bm{x}_t$ from $t = T$ to $t = 1$ as follows \cite{LDM}.
\begin{equation} 
    \bm{x}_{t-1}=\frac{1}{\sqrt{\alpha_t}}\left(\bm{x}_t-\frac{1-\alpha_t}{\sqrt{1-\bar{\alpha}_t}}\bm{\mu}_{\bm{\theta}}(\bm{x}_t,t)\right)+\beta_{t}\bm{I}.\label{ddpm-reverse}
\end{equation}
Remarkably, the second term in \eqref{ddpm-reverse} serves the purpose of obtaining the same distribution of $\bm{x}_{t-1}$ obtained through denoising, particularly in terms of variance, as that is derived through forward diffusion. If the second term of \eqref{ddpm-reverse} is removed, the $\bm{x}_{t-1}$ obtained by denoising is just equal to the mean of which obtained via forward diffusion procedure. Strictly, the scaling factor in the second term of \eqref{ddpm-reverse} should be $\frac{1-\bar{\alpha}_{t-1}}{1-\bar{\alpha}_{t}}\beta_{t}$ to ensure distribution consistency. However, as shown in \cite{ho2020denoising}, setting the scaling factor to $\beta_t$ achieves the same performance as setting it to $\frac{1-\bar{\alpha}_{t-1}}{1-\bar{\alpha}_{t}}\beta_{t}$, while also reducing computational complexity. Therefore, the scaling factor in the second term of \eqref{ddpm-reverse} is $\beta_t$ both for efficiency and effectiveness.

\section{Problem Formulation of RM Construction}\label{sec-3}
We consider the scenario where the RM needs to be constructed within an area as a grid of size $N \times N$, since the grid is small enough, the pathloss in a grid is a constant, where the RM can be represented by a matrix $\bm{P}$. Within this region, there is a BS with a single antenna and multiple static and dynamic obstacles. The location of BS can be represented by a tuple $\bm{R}$ as $\left<h,d_{x},d_{y}\right>$, where $h, d_{x}$ and $d_{y}$ is the height and coordinates of the BS. The static obstacles have varying sizes and shapes but are composed of the same surface material, thus reflecting and diffracting electromagnetic waves in the same way, and similar to \cite{zhang2023rme,levie2021radiounet,li2022radionet} the pathloss is zero within their interiors. The static obstacle information is represented by the matrix $\bm{H}_s \in \mathbb{R}^{N \times N}$, where a value of $h_{i,j}^{s} = 0,\forall h_{i,j}^{s} \in \bm{H}_s$ indicates the absence of the static obstacle at the position of $i-$th row and $j-$th column. Moreover, dynamic obstacles, such as vehicles, impact electromagnetic wave propagation through shielding, reflecting, and diffracting effects, similar to fixed obstacles, however, due to their small size and short height, dynamic obstacles cannot completely block the propagation of electromagnetic rays in its direction. The information of dynamic obstacles is represented by the matrix $\bm{H}_d \in \mathbb{R}^{N \times N}$, where the $h_{i,j}^{d} = 0,\forall h_{i,j}^{d} \in \bm{H}_d$ indicates the absence of dynamic obstacle at the position of $i-$th row and $j-$th column. 

The objective of this paper is to train an NN $\bm{\mu}_{\bm{\theta}}(\cdot)$ with parameters $\bm{\theta}$ to predict the pathloss matrix $\bm{\hat{P}} \in \mathbb{R}^{N \times N}$ based on the environmental features and location of the BS to minimize the difference between the predicted $\bm{\hat{P}}$ and the ground truth $\bm{P}$. The difference can be measured by the criterion function $\mathcal{L}(\hat{\bm{P}}, \bm{P})$. Therefore, the problem of RM construction can be formulated as
\begin{problem}\label{p1}
    \begin{align}
    &\min_{\bm{\theta}}&&\mathcal{L}(\hat{\bm{P}}, \bm{P}),\label{obj}\\
    &s.t. &&\hat{\bm{P}}=\bm{\mu}_{\bm{\theta}}(\bm{H}_{s}, \bm{H}_{d}, \bm{R})\tag{\ref{obj}a},
\end{align}
\end{problem}

The Problem \ref{p1} is a generative problem rather than a discriminative problem, which can be analyzed from both the data features and training method perspective. First, from the perspective of data features, the pathloss needs to be predicted from empty elements of the input data, as they do not exist in the environmental matrixes $\bm{H}_d$ and $\bm{H}_s$, which should be generated by the NN $\bm{\mu}_{\bm{\theta}}(\cdot)$. Moreover, since the pathloss to be predicted is not a discrete value, it is almost impossible to construct the RM by dividing the finite hyperplane to classify the nodes. However, from the perspective of statistical learning, the fundamental of discriminative tasks, especially supervised learning, is to let NN learn from the latent space to the partitioning of hyperplanes to achieve data classification \cite{vapnik1998statistical}. Thus, the discriminative method inevitably limits the performance of RM construction. Second, from the perspective of training methods, the self-supervised training method is mainly used by the generative model, where some elements of the raw data are masked and an NN is trained to recover the masked data through unmasked data in self-supervised learning \cite{goodfellow2016deep}. In the RM construction, the environmental information $\left<\bm{H}_{s}, \bm{H}_{d}\right>$ can be regarded as $\bm{P}$ whose pathloss elements are masked, and the NN needs to predict the masked pathloss elements based on unmasked data that are $\left<\bm{H}_{s}, \bm{H}_{d}\right>$. Therefore, the NN used for RM construction is trained in a self-supervised learning style. In addition, since the location of the BS $\bm{R}$ affects the distribution of pathloss, the BS can be regarded as a condition in self-supervision training, which means that the construction of RM is a condition generative problem.

\section{Diffusion-Based RM Construction}\label{sec-4}
As analyzed in section-\ref{sec-3} the RM construction is a generative problem, thus the SOTA generative diffusion model is used as a backbone to construct the RM effectively.

\subsection{Initial Processing}
To improve the convergence speed, the pathloss matrix $\bm{P}$ is encoded into a grayscale matrix through a process involving logarithmic scaling, normalization, and subsequent quantization \cite{levie2021radiounet}. Additionally, $\bm{R}$ is also represented as a grayscale matrix, with a pixel value of 1 denoting the location of the AP, while other pixels are set to 0. Subsequently, all the environment information is encoded into a three-channel tensor as a prompt tensor $\bm{C} = \left[\bm{H}_{s}, \bm{H}_{d}, \bm{R}\right]$. To further enhance the training efficiency of the denoise diffusion model, similar to \cite{LDM}, we initially train a variational autoencoder (VAE) to encode the raw data into the latent space for training and testing \cite{VAE}. The encoder $\mathcal{E}$ module of the VAE encodes $\bm{P}$ as a latent vector $\bm{z}_{0}$, where the noise is added to the $\bm{z}_{0}$ according to \eqref{diffused} for training. After the noise is removed by the NN, the decoder module $\mathcal{D}$ of the VAE is utilized to recover the RM from the diffusion prediction vector $\bm{\hat{z}}_{0}$. Through the use of the VAE, the denoise diffusion model only needs to remove the noise vector added to $\bm{z}$ instead of the noise matrix originally added to $\bm{P}$, thereby reducing the output space dimension of the diffusion model to enhance training efficiency. Thus, in the following of this paper, we use the latent vector $\bm{z}$ instead of $\bm{x}$ to denote the image to be generated. It is essential to highlight that the training of the VAE operates independently of the subsequent training of the denoise diffusion model, where the VAE only utilizes $\bm{P}$ for training the encoder $\mathcal{E}$ and decoder $\mathcal{D}$ trained in an autoencoder style. Then the VAE's parameters remain static and do not change in the subsequent training procedure of the diffusion model.

\begin{figure*}[t]
\captionsetup{font={small}, skip=16pt}
\centering
    \subfigure[The details of the framework.]
    {
     \centering
           \includegraphics[width=0.95\linewidth]{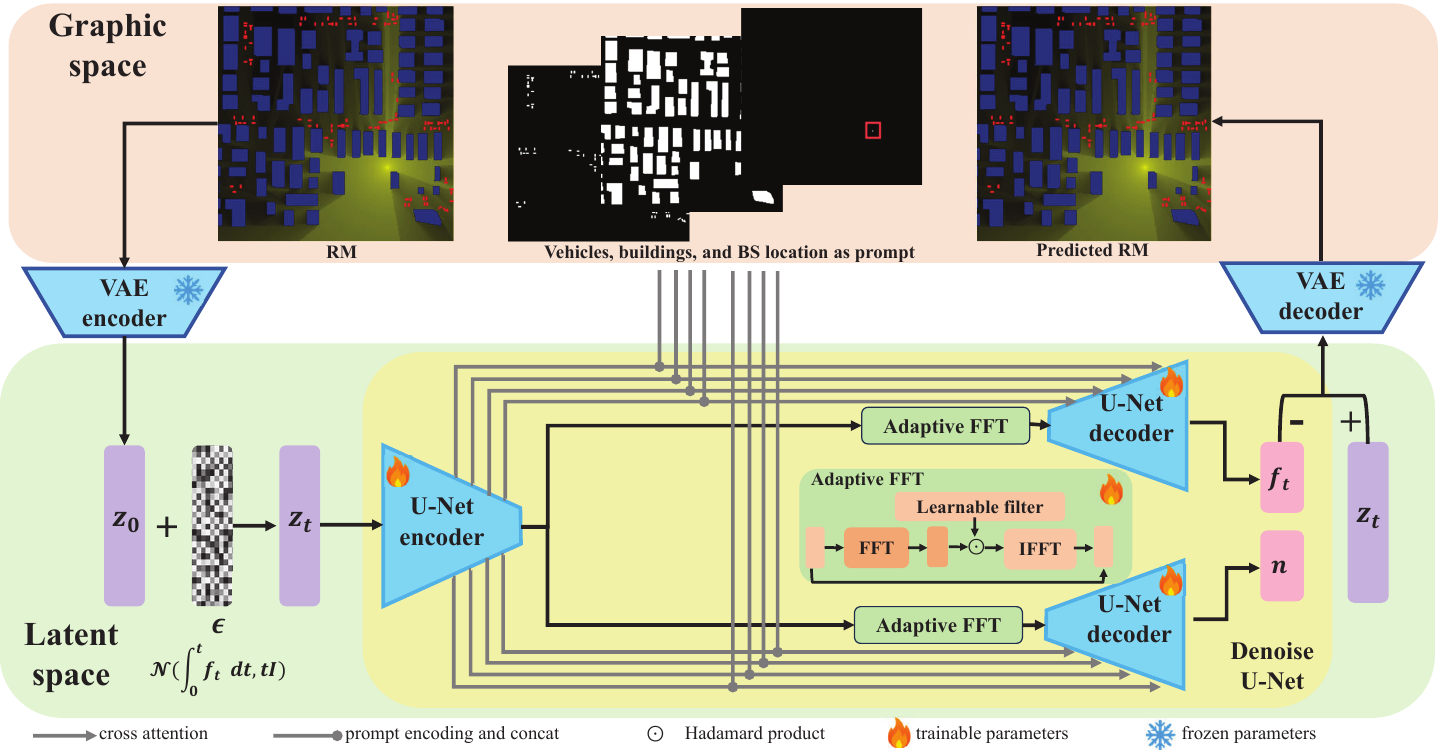}\label{fig_framwork}
    \vspace{-12pt}
    }
    \subfigure[The illustration of denoising procedure.]
    {
     \centering
           \includegraphics[width=0.95\linewidth]{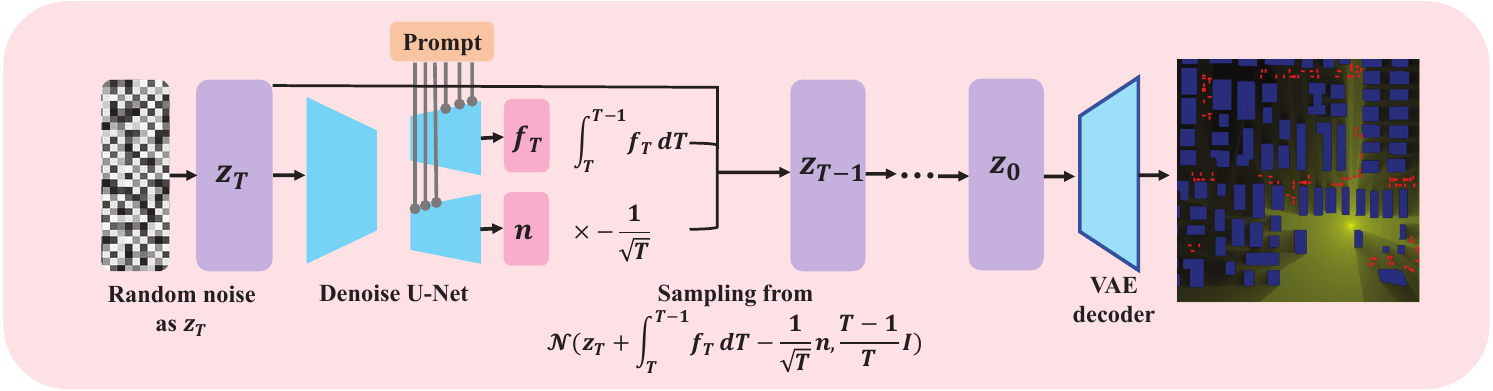}\label{fig_sampling}
    \vspace{-12pt}
    }
\vspace{-9pt}
\caption{The illustration of the proposed RadioDiff framework. The VAE is employed to encode the RM into a latent vector, thereby reducing the dimension of the input/output space for the denoise diffusion model. The framework incorporates a U-Net architecture, consisting of an encoder and decoder, to facilitate the denoising process. The prompt is represented as a grayscale diagram with three channels, each channel depicting the features of buildings, vehicles, and AP. After encoding the prompt, it is concatenated into the U-Net network, enabling the model to generate RMs under environmental conditions.
}
\vspace{-15pt}
\label{fig_method}
\end{figure*}

\subsection{RadioDiff via Decoupled Diffusion Model}
Although DDPM and latent diffusion models (LDM) have shown considerable potential across various fields, their extensive inference times and prolonged training durations motivate the use of a decoupled diffusion model (DDM) \cite{huang2024decoupled} for further enhancement. In DDM, the procedure of diffusing $\bm{z}_{0}$ to $\bm{z}_{t}$ is modeled as a two-stage continuous Markov process. First, $\bm{z}_{0}$ gradually diffuses into a $\bm{0}$ vector, then the noise $\bm{\epsilon}$ is added to this $\bm{0}$ vector to form $\bm{z}_{t}$. The distribution of $\bm{z}_t$ can be formulated as follows.
\begin{align}
    q\left(\bm{z}_t \mid \bm{z}_0\right)=\mathcal{N}\left(\gamma_t \bm{z}_0, \delta_t^2 \bm{I}\right),\label{ddm-diffuse}
\end{align}
where $\gamma_t$ and $\delta_t$ are hyper-parameters, and $\delta_t$ is designed to increase gradually over time while $\gamma_t$ decreases. 
\add{Compared to traditional diffusion models where the diffusion process directly adds noise to the original input $\bm{x}_0$ or $\bm{z}_0$, the decoupled diffusion model used in RadioDiff splits this process into two distinct stages. In DDM, $\bm{z}_0$ first diffuses into a $\bm{0}$ vector, which effectively decouples the contribution of the original input from the added noise. This allows for a more controlled diffusion process, as the noise $\bm{\epsilon}$ is only introduced after $\bm{z}_0$ has been reduced to a $\bm{0}$ vector. As a result, DDM reduces the variance in early diffusion steps and improves stability during both training and inference. Furthermore, this decoupled structure helps mitigate the prolonged inference time seen in traditional diffusion models, enabling more efficient generation in RadioDiff.}
Thus, according to \cite{song2020score}, \eqref{ddm-diffuse} can also be expressed as the following differential equation.
\begin{align}
    \mathrm{d} \bm{z}_t &= f_t \bm{z}_t \mathrm{~d} t + g_t \mathrm{~d} \bm{\epsilon}_t,\label{ddm-differential} \\
    f_t &= \frac{\mathrm{d} \log \gamma_t}{\mathrm{~d} t}, \\
    g_t^2 &= \frac{\mathrm{d} \delta_t^2}{\mathrm{~d} t} - 2 f_t \delta_t^2,
\end{align}
Based on the above equations, inverting $\bm{z}_t$ to $\bm{z}_0$ can be derived as follows.
\begin{align}
    \mathrm{d} \bm{z}_t = \left[f_t \bm{z}_t - g_t^2 \nabla_{\bm{x}} \log q\left(\bm{z}_t\right)\right] \mathrm{d} t + g_t \mathrm{d} \overline{\bm{\epsilon}}_t,\label{ddm-dx}
\end{align}
where $\overline{\bm{\epsilon}}_t$ is the Gaussian random variable in the reversed diffusion, playing a similar role as the second term of \eqref{ddpm-reverse}. By applying the decoupled diffusion strategy, the forward diffusion process can be redefined as follows.
\begin{align}
    &\bm{z}_t = \bm{z}_0 + \int_0^t \bm{f}_t \mathrm{d}t + \int_0^t \mathrm{d} \bm{\epsilon}_t,\label{ddm-add-noise} \\
    &\bm{z}_0 + \int_0^t \bm{f}_t \mathrm{d}t = \bm{0},\label{ddm-z-0}
\end{align}
where $\bm{z}_0 + \int_0^t \bm{f}_{t}\mathrm{d}t$ describes the data attenuation process and $\int_0^t d\bm{\epsilon}_t$ denotes the noise addition process. The function $\bm{f}_t$ is a differentiable function of $t$, and $\bm{\epsilon}_t$ is the standard Wiener process. Corresponding to equation \eqref{ddm-diffuse}, the conditional distribution can be simplified as follows.
\begin{equation}
    q(\bm{z}_t|\bm{z}_0) = \mathcal{N}\left(\bm{z}_{0}+\int_{0}^{t}\bm{f}_t\mathrm{d}t,t\bm{I}\right).
\end{equation}
By using the conditional probability formula the reverse sampling process, $\bm{z}_{t-\Delta t}$ can be obtained as follows.
\begin{align}
    q\left(\bm{z}_{t-\Delta t} \mid \bm{z}_t, \bm{z}_0\right)  &=\mathcal{N}\left(\bm{z}_{t} +\int_t^{t-\Delta t} \bm{f}_t \mathrm{~d} t-\frac{\Delta t}{\sqrt{t}} \bm{\epsilon}, \frac{\Delta t(t-\Delta t)}{t} \bm{I}\right).\label{reverse}
\end{align}
According to \eqref{reverse}, to reverse the noisy $\bm{z}_{t}$ into $\bm{z}_{0}$, which corresponds to the feature map of RM $\bm{P}$ in this paper, the NN only needs to predict two terms: $\int_t^{t-\Delta t} \bm{f}_t \mathrm{~d} t$ and $\bm{\epsilon}$, since $\bm{z}_{t}$ is a known vector to denoise. As shown in Fig.~\ref{fig_method}, two U-Net decoder networks are employed to predict $\int_t^{t-\Delta t} \bm{f}_t \mathrm{~d} t$ and $\bm{\epsilon}$, respectively. according to \eqref{ddm-add-noise}, the label for $\bm{\epsilon}$ is straightforward to obtain since the noise is added manually. However, the label for $\int_t^{t-\Delta t} \bm{f}_t \mathrm{~d} t$ must be obtained by solving the differential equation \eqref{ddm-z-0}. In the training procedure, the ground truth is obtained by solving $\bm{z}_0 + \int_0^1 \bm{f}_t \mathrm{~d} t = \bm{0}$. For a simple example where $\bm{f}_t$ is equal to a constant $\bm{c}$, the $\bm{f}_t$ can be easily determined as $\bm{f}_t = \bm{c} = -\bm{z}_0$. For another scenario where the value $\bm{f}_{t}$ is in linear dependence with $t$ that is $\bm{f}_t = \bm{a} t + \bm{b}$, solving the two parameters $\bm{a}$ and $\bm{b}$ using one equation is infeasible. To address this, we should sample one parameter from $\mathcal{N}(\bm{0}, \bm{I})$ and substitute it into $\bm{z}_0 + \int_0^1 \bm{f}_t \mathrm{~d} t = \bm{0}$ to solve the other parameter. In this way, we concatenate $\bm{a}$ and $\bm{b}$ to obtain the ground truth. The ground truths for other functions can be determined similarly. Since different $\bm{f}_t$ calculation methods have different effects on the data quality generated by diffusion, the coefficients of the equations for calculating ground truth of $\bm{f}_t$ can also be considered as hyper-parameters.

It should be emphasized that the above diffusion and denoising processes are based on constraint-free RM construction. However, according to Problem \ref{p1}, we need to achieve conditional RM construction based on environmental and BS location information, ensuring that the generated RM is correlated with these features. To enable the NN to generate the required RM based on the prompt $\bm{C}$, we employ a conditional generative model using an attention-based architecture. This approach leverages the attention mechanism, allowing the output of the NN to correlate with the given attention key-value vectors $\bm{K}$ and $\bm{V}$. Since the attention architecture cannot directly handle two-dimensional data, an extractor NN $\bm{\nu}(\cdot)$ projects the prompt $\bm{C}$ into an embedding space. This embedding is then projected to the intermediate layers of the U-Net through a cross-attention layer, implementing the attention mechanism $\text{Attention}(\bm{Q}, \bm{K}, \bm{V}) = \text{softmax}\left(\frac{\bm{QK}^T}{\sqrt{d}}\right) \cdot \bm{V}$ as follows.
\begin{align}
    &\bm{Q} = \bm{W_{Q}} \cdot \bm{\phi}(\bm{z}_{t}), \\
    &\bm{K} = \bm{W_{K}} \cdot \bm{\nu}(\bm{C}), \\
    &\bm{V} = \bm{W_{V}} \cdot \bm{\nu}(\bm{C}),
\end{align}
where $\bm{\phi}(\bm{z}_t)$ is the flattened operator applied on the output from the U-Net, which executes the function $\theta$ and applies the transformation $\bm{W_{V}}$. Additionally, $\bm{W_{Q}}$ and $\bm{W_{K}}$ represent trainable projection matrices \cite{vaswani2017attention}. The training loss function can be represented as follows.
\begin{align}
    &\mathcal{L} = \mathbb{E}_{z, c, t, \epsilon} \left[ \left\| \bm{\epsilon} - \bm{\epsilon}_{\theta} \right\|^2 + \left\| \bm{f} - \bm{f}_{\bm{\theta}} \right\|_2^2 \right], \\
    &\bm{\epsilon}_{\theta}, \bm{f}_{\bm{\theta}} = \bm{\mu}_{\bm{\theta}}\left(\bm{z}_{t}; \bm{C}, t \right).
\end{align}

\subsection{Adaptive FFT Filter for DRM Enhancement}
As shown in Fig.~\ref{fig-example}, RM \add{exhibits numerous edge texture features, especially in DRM, which generate substantial high-frequency information in the frequency domain} \cite{chen2024frequency}. Although conventional convolutional layers within the denoising U-Net \add{effectively extract features, they struggle to precisely capture high-frequency components}. \add{As a result, neural networks (NNs) based solely on traditional convolutional layers often produce overly smooth RM outputs, leading to blurred representations and suboptimal performance in dynamic environments.}

\add{To address this issue, as depicted in Fig.~\ref{fig_method}, an adaptive Fast Fourier Transform (FFT) filter (AFT) module is introduced, specifically to enhance the model’s capacity for extracting high-frequency features.} The AFT module \add{operates by transforming the 2D feature maps} $\bm{z}$, \add{generated by the encoder with spatial dimensions $H \times W$ and channel count $C$, from the spatial domain to the frequency domain via the FFT operation, represented as} $\bm{z}_{c} = \mathcal{F}(\bm{z})$, where $\mathcal{F}$ denotes the FFT. \add{To improve the model's ability to focus on relevant frequency components,} AFT \add{incorporates a learnable weight matrix} $\bm{w}\in \mathbb{C}^{H \times W \times C}$. This matrix is applied to the frequency-domain features $\bm{z}_{c}$ via the Hadamard product, $\bm{w}\odot \bm{z}_{c}$. The \add{learnable weight matrix dynamically adjusts the model's response to different frequency distributions in the target data, emphasizing important frequencies while attenuating irrelevant ones}. This \add{adaptive spectral filtering allows the model to perform global frequency adjustments based on data-driven relevance, ensuring that high-frequency features are enhanced while less critical frequencies are suppressed}.

Following this adaptive filtering process, the modified frequency-domain features are transformed back into the spatial domain using the Inverse Fast Fourier Transform (IFFT). \add{To preserve key information and mitigate potential losses during filtering, a residual connection is introduced between the original feature map} $\bm{z}$ \add{and the output feature map}. The complete operation of the AFT module can be described as follows:

\begin{equation}
    \bm{z} = \bm{z} + \mathcal{F}^{-1}(\bm{w}\odot \bm{z}_{c}),
\end{equation}
\begin{equation}
    \bm{z}_{c} =\mathcal{F}(\bm{z}).
\end{equation}

\section{Experiments}\label{sec-5}
\subsection{Datasets}
In this study, we evaluate the performance of the proposed method using the RadioMapSeer dataset provided by the pathloss RM construction challenge \cite{yapar2023first}. 
\add{The dataset consists of 700 maps, each with unique geographic information (e.g., building data), with each map containing 80 transmitter locations and their corresponding ground truth data. Each map contains between 50 and 150 buildings. We selected 500 maps for the training dataset and the remaining 200 maps for the test dataset. There is no overlapping terrain information between the training and test datasets.} 

\add{The city maps include data from cities such as Ankara, Berlin, Glasgow, Ljubljana, London, and Tel Aviv, sourced from OpenStreetMap. In the dataset, both transmitter and receiver heights are set at 1.5 meters, while building heights are set at 25 meters. Each map is converted into a 256 × 256 pixel morphological 2D image with binary pixel values (0/1), where each pixel represents one square meter: ‘1’ for areas inside buildings and ‘0’ for areas outside.}
Transmitter positions are stored in a two-dimensional numerical format and depicted in morphological images, with the transmitter's pixel set to `1' and all others to `0'. The transmitter power is set to 23 dBm, and the carrier frequency is 5.9 GHz. To obtain a sufficiently accurate RM as ground truth for training, the ground truth RM in the dataset is constructed using Maxwell's equations, where the pathloss is calculated by the reflection and diffraction of electromagnetic rays. The RMs that only consider the impact of the static buildings on the electromagnetic rays are used as the ground truth of SRM. Additionally, the RMs in the dataset that consider the impact both of the static buildings and the vehicles, which are randomly generated along the roads, are used as the ground truth of the DRM, as shown in Fig.~\ref{fig-example}.
 \begin{figure*}[t]
\captionsetup{font={small}, skip=16pt}
\scriptsize
%\centering
\begin{tabular}{ccc}
% one row
\hspace{-0.5cm}
\begin{adjustbox}{valign=t}
\begin{tabular}{c}
\end{tabular}
\end{adjustbox}
\begin{adjustbox}{valign=t}
\begin{tabular}{cccccccc}
\includegraphics[width=0.194\linewidth]{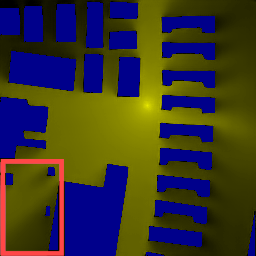} \hspace{-4mm} &
\includegraphics[width=0.194\linewidth]{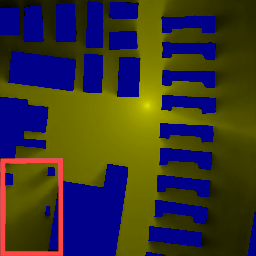}   \hspace{-4mm} &
\includegraphics[width=0.194\linewidth]{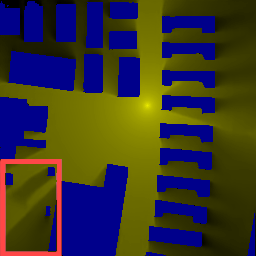}  \hspace{-4mm} &
\includegraphics[width=0.194\linewidth]{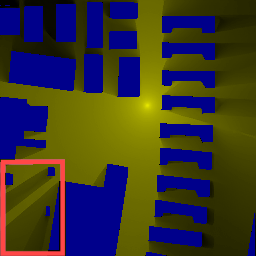}  \hspace{-4mm} &
\includegraphics[width=0.194\linewidth]{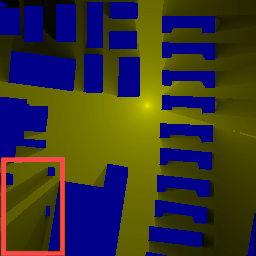}  \hspace{-4mm} &
\end{tabular}
\end{adjustbox}
\vspace{0.1mm}
\\
\hspace{-0.5cm}
\begin{adjustbox}{valign=t}
\begin{tabular}{c}
\end{tabular}
\end{adjustbox}
\begin{adjustbox}{valign=t}
\begin{tabular}{cccccccc}
\includegraphics[width=0.194\linewidth]{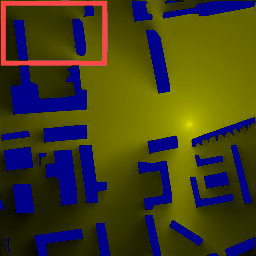} \hspace{-4mm} &
\includegraphics[width=0.194\linewidth]{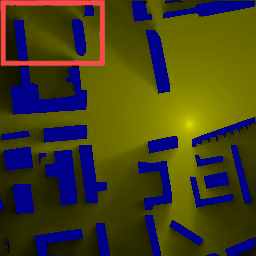}   \hspace{-4mm} &
\includegraphics[width=0.194\linewidth]{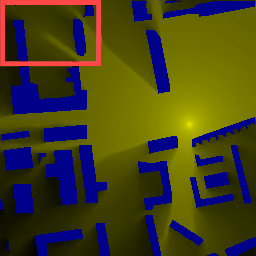}  \hspace{-4mm} &
\includegraphics[width=0.194\linewidth]{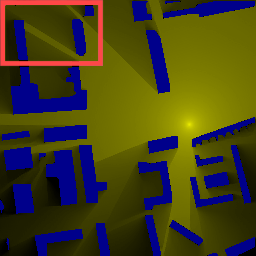}  \hspace{-4mm} &
\includegraphics[width=0.194\linewidth]{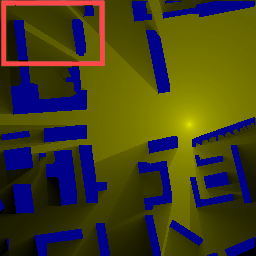}  \hspace{-4mm} &
\end{tabular}
\end{adjustbox}
\vspace{0.1mm}
\\
\hspace{-0.5cm}
\begin{adjustbox}{valign=t}
\begin{tabular}{c}
\end{tabular}
\end{adjustbox}
\begin{adjustbox}{valign=t}
\begin{tabular}{cccccccc}
\includegraphics[width=0.194\linewidth]{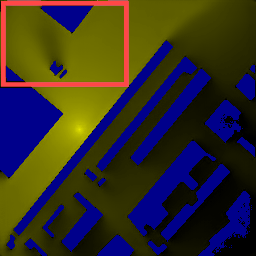}  \hspace{-4mm} &
\includegraphics[width=0.194\linewidth]{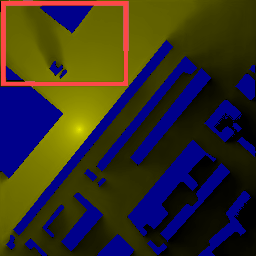}  \hspace{-4mm} &
\includegraphics[width=0.194\linewidth]{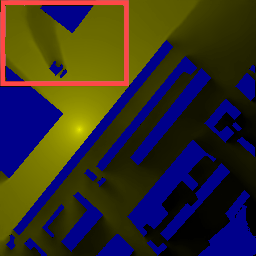}  \hspace{-4mm} &
\includegraphics[width=0.194\linewidth]{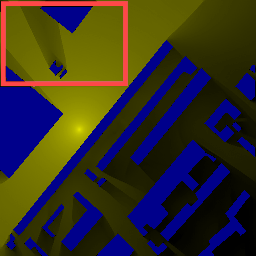} \hspace{-4mm} &
\includegraphics[width=0.194\linewidth]{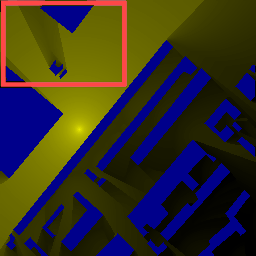}  \hspace{-4mm} &
\end{tabular}
\end{adjustbox}
\vspace{0.1mm}
\\
% one row
\hspace{-0.55cm}
\begin{adjustbox}{valign=t}
\begin{tabular}{c}
\end{tabular}
\end{adjustbox}
\begin{adjustbox}{valign=t}
\begin{tabular}{cccccccc}
\includegraphics[width=0.194\linewidth]{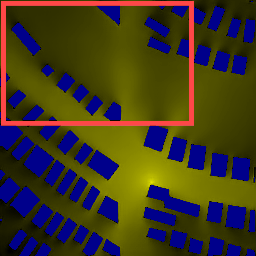}  \hspace{-4mm} &
\includegraphics[width=0.194\linewidth]{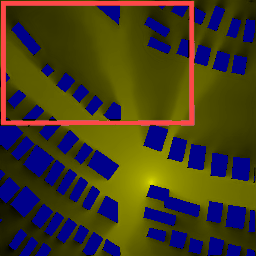}  \hspace{-4mm} &
\includegraphics[width=0.194\linewidth]{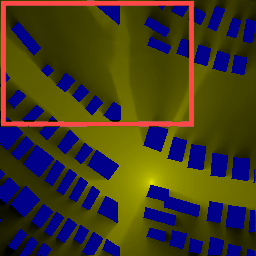}  \hspace{-4mm} &
\includegraphics[width=0.194\linewidth]{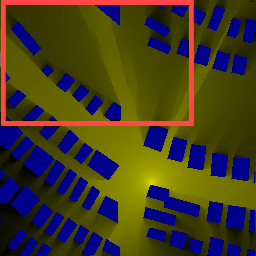} \hspace{-4mm} &
\includegraphics[width=0.194\linewidth]{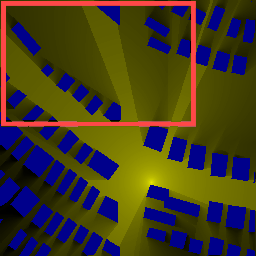}  \hspace{-4mm} &
\\
RME-GAN \hspace{-4mm} &
UVM-Net \hspace{-4mm} &
RadioUNet \hspace{-4mm} &
RadioDiff (Ours) \hspace{-4mm} &
Ground Truth
\\
\end{tabular}
\end{adjustbox}
\end{tabular}
% \vspace{-5.mm}
\caption{The comparisons of constructed SRM on different methods.}
\label{fig:r}
\vspace{-12pt}
\end{figure*}

\subsection{Metrics}
% To comprehensively evaluate the quality of RM constructions, we utilize MSE, structural similarity index measurement (SSIM), and peak signal-to-noise ratio (PSNR) simultaneously as metrics, where MSE quantifies the overall error that is the accuracy of the RM construction, SSIM evaluates the preservation of structural information and detail precision, and PSNR measures the ratio of signal to noise affecting the fidelity of the constructed RM.

\add{To comprehensively evaluate the quality of RM construction, we begin by adopting the parameters commonly used in previous studies \cite{levie2021radiounet}, namely NMSE and RMSE. Additionally, we observe that the accurate generation of structural information and details is a key objective in RM reconstruction tasks, whereas the MSE index focuses on overall error and does not directly address these specific requirements. Therefore, we propose introducing structural similarity index measurement (SSIM) and peak signal-to-noise ratio (PSNR) as additional metrics in this paper. SSIM evaluates the preservation of structural information to emphasize the accuracy of structural detail reconstruction, while PSNR measures the signal-to-noise ratio to assess the fidelity of RM construction, particularly with respect to edge signal reconstruction.}

\subsubsection{MSE} MSE is calculated by averaging the squared differences between the pixel intensities of the original and final images as follows.
\begin{equation}MSE=\frac1{NM}\Sigma_{m=0}^{M-1}\sum_{n=0}^{N-1}e(m,n)^2,\end{equation}
where $e(m,n)$ is the error difference between the ground truth and the predicted RM, and $M,N$ is the length and width of the image, respectively. The normalized MSE (NMSE) is a scaled version of MSE employed to assess the predictive accuracy of the RM construction, where rooted MSE (RMSE) is the rooted MSE, which are defined as follows.
\begin{equation}NMSE=\frac{\Sigma_{m=1}^M\Sigma_{n=1}^N(I_b(m,n)-I(m,n))^2}{\Sigma_{m=1}^M\sum_{n=1}^NI^2(m,n)},\end{equation}
\begin{equation}\mathrm{RMSE}=\sqrt{MSE}\end{equation}

\begin{table}[t]
\captionsetup{font={small}, skip=14pt}
  \centering
  \caption{\textbf{Quantitative Comparison.} Results in bold red and underlined blue highlight the highest and second highest, respectively. The $\uparrow$ indicates metrics whereby higher values constitute improved outcomes, with higher values preferred for all other metrics.}
  \vspace{-6pt}
\resizebox{0.5\linewidth}{!}{
\begin{tabular}{@{}cc|cccc@{}}
\toprule
\multicolumn{2}{c|}{Methods} & RME-GAN & RadioUNet & UVM-Net & RadioDiff (Ours) \\ \midrule
 & NMSE & 0.0115 & {\color[HTML]{00009B} \underline{0.0074}} & 0.0085 & {\color[HTML]{9A0000} \textbf{0.0049}} \\
 & RMSE & 0.0303 & {\color[HTML]{00009B} \underline{0.0244}} & 0.0304 & {\color[HTML]{9A0000} \textbf{0.0190}} \\
 & SSIM $\uparrow$& 0.9323 & {\color[HTML]{00009B} \underline{0.9592}} & 0.9320 & {\color[HTML]{9A0000} \textbf{0.9691}} \\
\multirow{-4}{*}{SRM} & PSNR $\uparrow$ & 30.54 & {\color[HTML]{00009B} \underline{32.01}} & 30.34 & {\color[HTML]{9A0000} \textbf{35.13}} \\ \midrule
 & NMSE & 0.0118 & {\color[HTML]{000000} 0.0089} & {\color[HTML]{00009B} \underline{0.0088}} & {\color[HTML]{9A0000} \textbf{0.0057}} \\
 & RMSE & 0.0307 & {\color[HTML]{00009B} \underline{0.0258}} & 0.0301 & {\color[HTML]{9A0000} \textbf{0.0215}} \\
 & SSIM $\uparrow$& 0.9219 & {\color[HTML]{00009B} \underline{0.9410}} & 0.9326 & {\color[HTML]{9A0000} \textbf{0.9536}} \\
\multirow{-4}{*}{DRM} & PSNR $\uparrow$& 30.40 & {\color[HTML]{00009B} \underline{31.75}} & 30.42 & {\color[HTML]{9A0000} \textbf{34.92}} \\ \bottomrule
\end{tabular}
}
\vspace{-2mm}

  \label{tab1}
\vspace{-12pt}
\end{table}

 \begin{figure*}[t]
\captionsetup{font={small}, skip=16pt}
\scriptsize
%\centering
\begin{tabular}{ccc}
% one row
\hspace{-0.5cm}
\begin{adjustbox}{valign=t}
\begin{tabular}{c}
\end{tabular}
\end{adjustbox}
\begin{adjustbox}{valign=t}
\begin{tabular}{cccccccc}
\includegraphics[width=0.194\linewidth]{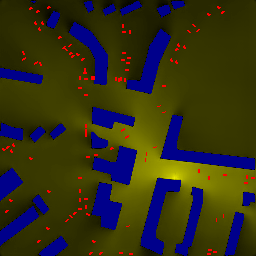} \hspace{-4mm} &
\includegraphics[width=0.194\linewidth]{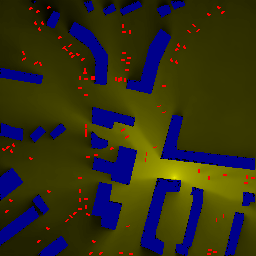}   \hspace{-4mm} &
\includegraphics[width=0.194\linewidth]{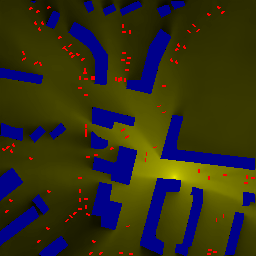}  \hspace{-4mm} &
\includegraphics[width=0.194\linewidth]{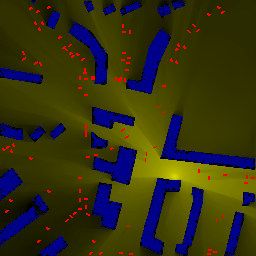}  \hspace{-4mm} &
\includegraphics[width=0.194\linewidth]{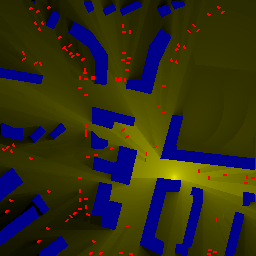}  \hspace{-4mm} &
\end{tabular}
\end{adjustbox}
\vspace{0.1mm}
\\
\hspace{-0.5cm}
\begin{adjustbox}{valign=t}
\begin{tabular}{c}
\end{tabular}
\end{adjustbox}
\begin{adjustbox}{valign=t}
\begin{tabular}{cccccccc}
\includegraphics[width=0.194\linewidth]{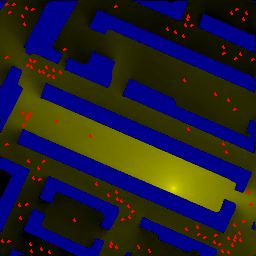} \hspace{-4mm} &
\includegraphics[width=0.194\linewidth]{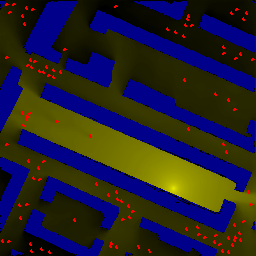}   \hspace{-4mm} &
\includegraphics[width=0.194\linewidth]{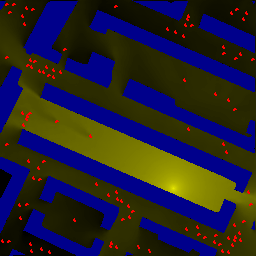}  \hspace{-4mm} &
\includegraphics[width=0.194\linewidth]{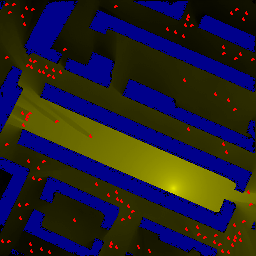}  \hspace{-4mm} &
\includegraphics[width=0.194\linewidth]{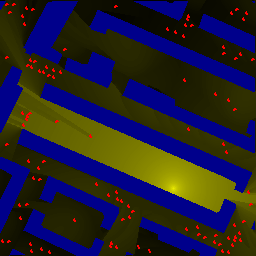}  \hspace{-4mm} &
\end{tabular}
\end{adjustbox}
\vspace{0.1mm}
\\
\hspace{-0.5cm}
\begin{adjustbox}{valign=t}
\begin{tabular}{c}
\end{tabular}
\end{adjustbox}
\begin{adjustbox}{valign=t}
\begin{tabular}{cccccccc}
\includegraphics[width=0.194\linewidth]{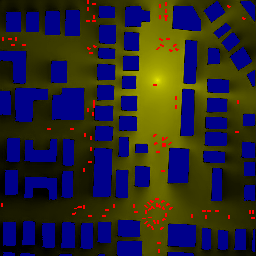} \hspace{-4mm} &
\includegraphics[width=0.194\linewidth]{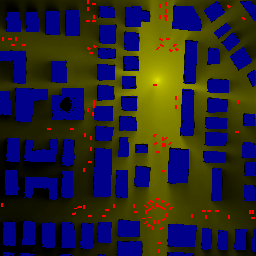}   \hspace{-4mm} &
\includegraphics[width=0.194\linewidth]{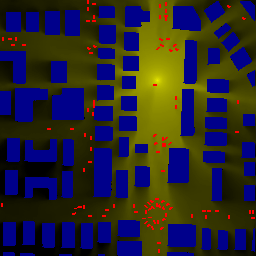}  \hspace{-4mm} &
\includegraphics[width=0.194\linewidth]{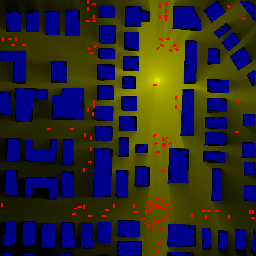}  \hspace{-4mm} &
\includegraphics[width=0.194\linewidth]{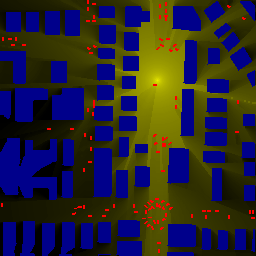}  \hspace{-4mm} &
\end{tabular}
\end{adjustbox}
\vspace{0.1mm}
\\
% one row
\hspace{-0.55cm}
\begin{adjustbox}{valign=t}
\begin{tabular}{c}
\end{tabular}
\end{adjustbox}
\begin{adjustbox}{valign=t}
\begin{tabular}{cccccccc}
\includegraphics[width=0.194\linewidth]{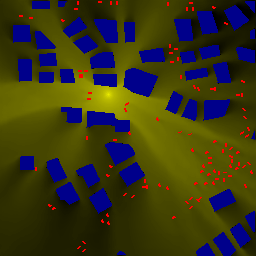} \hspace{-4mm} &
\includegraphics[width=0.194\linewidth]{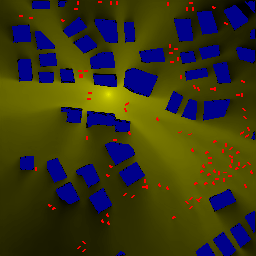}   \hspace{-4mm} &
\includegraphics[width=0.194\linewidth]{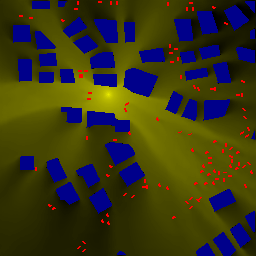}  \hspace{-4mm} &
\includegraphics[width=0.194\linewidth]{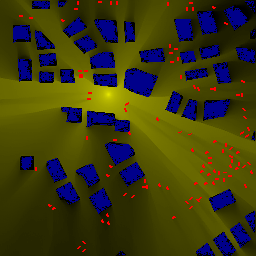}  \hspace{-4mm} &
\includegraphics[width=0.194\linewidth]{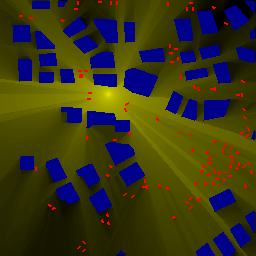}  \hspace{-4mm} &
\\
RME-GAN \hspace{-4mm} &
UVM-Net \hspace{-4mm} &
RadioUNet \hspace{-4mm} &
RadioDiff (Ours) \hspace{-4mm} &
Ground Truth
\\
\end{tabular}
\end{adjustbox}
\end{tabular}
% \vspace{-5.mm}
\caption{The comparisons of constructed DRM on different methods.}
\label{fig:r1}
\vspace{-12pt}
\end{figure*}

\subsubsection{SSIM}
SSIM is a quality assessment metric inspired by the human visual system. Since SSIM focuses on measuring texture differences, and there are lots of high-frequency details in RM, SSIM is suitable for evaluating the quality of the generated results. \add{We also believe that greater attention should be given to the brightness of the signal radiation, the contrast between the signal radiation and the surrounding area, and the accuracy of geographic map in RM reconstruction. This aligns with the SSIM metric, which evaluates three key components: brightness, contrast, and structural information}, which can be calculated as follows.
\begin{equation}l(x,y)=\frac{2\mu_X(x,y)\mu_Y(x,y)+C_1}{\mu_X^2(x,y)+\mu_Y^2(x,y)+C_1}\end{equation}
\begin{equation}c(x,y)=\frac{2\sigma_X(x,y)\sigma_Y(x,y)+C_2}{\sigma_X^2(x,y)+\sigma_Y^2(x,y)+C_2}\end{equation}
\begin{equation}s(x,y)=\frac{\sigma_{XY}(x,y)+C_3}{\sigma_X(x,y)\sigma_Y(x,y)+C_3}\end{equation}
where $x$, $y$ correspond to two different input images and $\mu_x, \sigma_x^2, \sigma_{xy}$ denote the mean and variance of $x$ and the covariance of $x$ and
$y$ respectively. In addition, $ C_1 $, $ C_2 $, and $ C_3 $ are constants which are defined as follows.
$$
C_1 = (K_1L)^2, \quad C_2 = (K_2L)^2, \quad C_3 = \frac{C_2}{2},
$$
where $ L $ represents the dynamic range of the data. Based on these parameters, the structural similarity can be computed as described as follows.
\begin{equation}SSIM(x,y)=\frac{(2\mu_x\mu_y+c_1)(\sigma_{xy}+c_2)}{(\mu_x^2+\mu_y^2+c_1)(\sigma_x^2+\sigma_y^2+c_2)}\end{equation}
\subsubsection{PSNR}
The PSNR is defined as the ratio between the maximum possible power of a signal and the power of interfering noise that affects the fidelity of its representation. PSNR is typically expressed in decibels (dB) and provides an approximate measure of the perceived quality of reconstruction. In image evaluation, a higher PSNR generally indicates better image quality. For RMs, an accurate edge signal is crucial; therefore, PSNR is used not only to assess overall image quality but also to determine the quality of edge detail in the generated RMs. PSNR can be calculated as follows.
\begin{equation}PSNR=10\log_{10}\left(\frac{r^2}{MSE}\right)\end{equation}
where $r$ is the maximal variation in the input image data.

\subsection{Implementation}
We implemented our RadioDiff framework using the PyTorch framework. The training process is divided into two phases, both utilizing an AdamW optimizer with a decaying learning rate, starting from $5 \times 10^{-5}$ and reducing to $5 \times 10^{-6}$. In the initial phase, the autoencoder is trained using RM images from the entire dataset's training set as ground truth. \add{This phase, which trains the VAE with $z$-channels set to 3 and embedding dimension of 128, takes approximately 120 hours on $4\times$ NVIDIA A100 SXM GPUs with a batch size of 2.} The subsequent phase involves training the denoise diffusion U-net model, which takes around 360 hours using $4\times$ NVIDIA A100 SXM GPUs with a batch size of 64. 
\add{In this implementation, the input image size is $256 \times 256$. The diffusion process uses $T = 1000$ in training, and the loss function is $l2$-based with an objective of predicting KC. The start distribution is set to normal, and the perceptual weight is set to $0$. }
Although a larger batch size can achieve better training results and higher training speed \cite{goodfellow2016deep}, the batch size in the first stage is significantly smaller than that in the second stage. In the first stage, both the input and output are raw image data, which consumes substantial video memory. In contrast, during the second stage, when training the diffusion model, the data is processed in the latent space encoded by the VAE, thereby reducing video memory consumption. Consequently, a larger batch size can be employed to enhance training speed and accuracy. This highlights the importance of using the VAE to encode data into the latent space. 
\add{The hyper-parameter for the diffusion timesteps $T$ is set to 500 for inference, and the diffusion process is decoupled with $\bm{f}_t$ defined as $-\bm{z}_0$.}

% We implemented our RadioDiff framework using the PyTorch framework. The training process is divided into two phases, both utilizing an AdamW optimizer with a decaying learning rate, starting from $5 \times 10^{-5}$ and reducing to $5 \times 10^{-6}$. In the initial phase, the autoencoder is trained using RM images from the entire dataset's training set as ground truth. This phase, which trains the VAE, takes approximately 120 hours on $4\times$ NVIDIA A100 SXM GPUs with a batch size of 2. The subsequent phase involves training the denoise diffusion model, which takes around 360 hours using $4\times$ NVIDIA A100 SXM GPUs with a batch size of 64. Although a larger batch size can achieve better training results and higher training speed \cite{goodfellow2016deep}, the batch size in the first stage is significantly smaller than that in the second stage. In the first stage, both the input and output are raw image data, which consumes substantial video memory. In contrast, during the second stage, when training the diffusion model, the data is processed in the latent space encoded by the VAE, thereby reducing video memory consumption. Consequently, a larger batch size can be employed to enhance training speed and accuracy. This highlights the importance of using the VAE to encode data into the latent space. For hyper-parameters, the $T$ is $500$, and $\bm{f}_{t}$ is $-\bm{z}_{0}$.

\subsection{Comparisons with SOTA Methods}
To evaluate the proposed RadioDiff model, we compare it with other SOTA methods. 
\add{To ensure a comprehensive comparison of the experiments, we compare the CNN-based, GAN-based, and Mamba-based methods separately, which represent the primary architectures used in the current RM reconstruction task based on deep learning.}
\add{For the detailed parameter settings of the comparison model, we adhere to the description provided in the article \cite{levie2021radiounet,zheng2024u,zhang2023rme}. Additionally, for fairness, the training and test data will be aligned with RadioDiff.
}
The following method is used to compare.
\begin{itemize}
    \item RadioUNet \cite{levie2021radiounet}: RadioUNet is one of the most effective sampling-free NN-based RM construction method, where a convolutional U-Net is used as the backbone NN, and the supervised learning is used to train the RadioUNet. \add{RadioUNet reconstructs the wireless propagation graph using a simple yet effective network architecture that learns environmental characteristics, making it one of the most representative reconfiguration algorithms based on reel machines.}
    
    \item UVM-Net \cite{zheng2024u}: The training method and settings of UVM-Net are same as RadioUNet, \add{but the backbone network is replaced by a convolutional layer with the latest SSMs (State Space Sequence Models). SSMs are specifically designed to handle long-sequence data and are particularly well-suited for modeling long-range dependencies. SSMs map the input sequence to a hidden state through a state space model and predict the output based on this hidden state. As a result, SSMs exhibit enhanced local feature capture and efficient remote modeling capabilities. We selected UVM-Net as the baseline to evaluate the performance differences between this SSM-driven architecture and traditional convolutional networks in the wireless propagation graph reconstruction task.}
    
    \item RME-GAN \cite{zhang2023rme}: The SOTA NN-based RM construction method, which uses a generative model cGAN to construct RM. However, RME-GAN is sampling-based, it not only uses environmental features but also uses the sampling pathloss as the input to construct the RM. For a fair comparison, the RME-GAN in this paper only uses environmental features as input. \add{As GAN models have been at the forefront of generative model research in recent years, they effectively showcase the unique capabilities and challenges of generative adversarial networks in handling RM reconstruction tasks.}
    
\end{itemize}

\subsubsection{Comparisons for SRM}
For the quantitative comparison on the RadioMapSeer-Test dataset for SRM scenarios is given in the first part of Table \ref{tab1} and Fig.~\ref{fig:r}, our model outperforms others in error metrics, i.e., NMSE, RMSE, and structural metrics, i.e., SSIM, PSNR, indicating that our predictions and generated RM are more accurate. Notably, RadioDiff excels in the PSNR metric, indicating that the RMs it generates have clearer and sharper structural edges compared to other methods. Furthermore, the qualitative comparison presented in Fig.~\ref{fig:r} demonstrates that the RMs constructed by RadioDiff closely resemble the ground truth, with well-defined edge features. This precision stems from the diffusion model's heightened sensitivity to the high-frequency signals of edge information, while the AFTs effectively enhance and isolate these signals. In contrast, structures such as CNNs and Mamba are less responsive to edge information, as is shown in RadioUNet and UVM-Net since they show inaccuracies in radiation signal positioning and edge blurring. As for the generative model RME-GAN, which relies on adversarial strategies, it tends to yield ambiguous and less precise results since it relies on the sampling position measurements, so in the sampling-free RM construction scenario it performs poorly.
\subsubsection{Comparisons for DRM}
As shown in the second part of Table \ref{tab1} and Fig.~\ref{fig:r1}, the quantitative comparison on the RadioMapSeer-Test dataset for DRM scenarios is given. In DRM scenarios, the models must account for additional dynamic environmental factors. Despite a general decline in performance, the second part of Table \ref{tab1} shows that RadioDiff consistently delivers the best results across all indicators. As shown in Fig.~\ref{fig:r1}, the RadioDiff model exhibits enhanced sensitivity to dynamic environmental factors such as vehicles, whereas the RME-GAN, RadioUNet, and UVM-Net models struggle with these elements, often resulting in significant blurring and distortion. This further highlights that RadioDiff has stable high performance under more challenging conditions, particularly in scenarios characterized by complex environments and overlapping signals.

\begin{figure}[t]
\captionsetup{font={small}, skip=8pt}
\scriptsize
\centering
\begin{tabular}{ccc}
% one row
\hspace{-0.5cm}
\begin{adjustbox}{valign=t}
\begin{tabular}{c}
\end{tabular}
\end{adjustbox}
\begin{adjustbox}{valign=t}
\begin{tabular}{cccccc}
\includegraphics[width=0.32\linewidth]{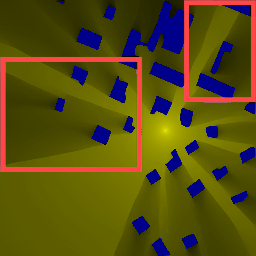} \hspace{-4mm} &
\includegraphics[width=0.32\linewidth]{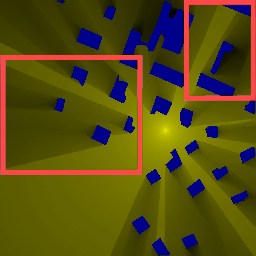} \hspace{-4mm} &
\includegraphics[width=0.32\linewidth]{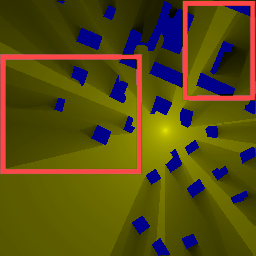} \hspace{-4mm} &
\\
w/o AFT \hspace{-4mm} &
w/ AFT \hspace{-4mm} &
Ground Truth \hspace{-4mm} &
\\

\end{tabular}
\end{adjustbox}
\vspace{0.5mm}
\\
% one row
\hspace{-0.55cm}
\begin{adjustbox}{valign=t}
\begin{tabular}{c}
\end{tabular}
\end{adjustbox}
\begin{adjustbox}{valign=t}
\begin{tabular}{cccccc}
\includegraphics[width=0.32\linewidth]{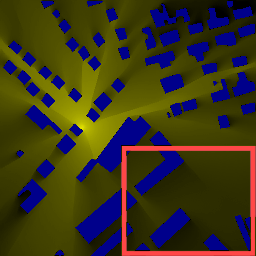} \hspace{-4mm} &
\includegraphics[width=0.32\linewidth]{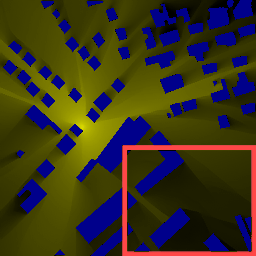} \hspace{-4mm} &
\includegraphics[width=0.32\linewidth]{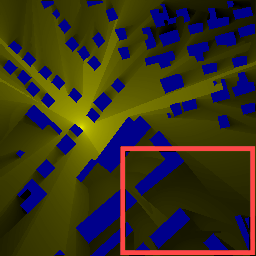} 
\hspace{-4mm} &
\\
w/o AFT \hspace{-4mm} &
w/ AFT \hspace{-4mm} &
Ground Truth \hspace{-4mm} &
\\

\end{tabular}
\end{adjustbox}
\end{tabular}
\caption{Ablation study about AFT. 
The qualitative results demonstrate that incorporating the AFT further enhances the model's sensitivity to edge signals. This leads to RM images with more accurate edges and more robust results when multiple signals overlap.}
\vspace{-9pt}
\label{fig:fft}
\end{figure}

\begin{table}[t]
\centering
\caption{Ablation Study about AFT.}
\vspace{-4pt}
\captionsetup{font={small}, skip=8pt}
\resizebox{0.5\linewidth}{!}{
\begin{tabular}{@{}c|cccc@{}}
\toprule
\textbf{\quad RadioDiff \quad} & \textbf{\,NMSE\,} & \textbf{\,RMSE\,} & \textbf{\,PSNR\,} & \textbf{\,SSIM\,} \\ \midrule
w/o AFT & 0.0067 & 0.0259 & 31.62 & 0.9465 \\
w/ AFT & {\color[HTML]{9A0000} \textbf{0.0049}} & {\color[HTML]{9A0000} \textbf{0.0190}} & {\color[HTML]{9A0000} \textbf{35.13}} & {\color[HTML]{9A0000} \textbf{0.9691}} \\ \bottomrule
\end{tabular}
}
\vspace{-12pt}
\label{tab2}
\end{table}

\subsection{Ablation Study}
In this section, we analyze the impact of the AFT on the performance of the RadioDiff model. The qualitative results in Fig.~\ref{fig:fft} vividly illustrate the visual disparity between radio patterns generated with and without employing the AFT. By incorporating the AFT, our model exhibits an enhanced capability to accurately detect and represent edge signals. This improvement is particularly noticeable when dealing with scenarios involving signal superposition, as evidenced by images produced by our model equipped with AFTs that possess increasingly sharper edges. Furthermore, Table \ref{tab2} presents quantitative comparison results, demonstrating that a better performance can be achieved through the utilization of the AFT.

\subsection{Limitations and Discussion}
Although this paper proposes a pioneer exploration of utilizing diffusion models for RM construction, and the proposed RadioDiff achieves SOTA performance, the issue of efficiency remains a key consideration for such a large generative model-based method. Table \ref{tab3} presents a comparison of the inference time and memory usage between the RadioDiff model and alternative models, 
\add{showing that the diffusion model requires more resources and has a longer inference time than other methods. However, it should be emphasized that although the proposed RadioDiff is more time-consuming than other NN-based methods, the inference delay is still less than one second, which remains acceptable for dynamically constructing RM. For training data, RadioDiff uses the same dataset as other NN-based methods, and no limitations regarding its access to training data were identified in the experiment. In addition, although VAE needs to be trained separately before RadioDiff’s formal training, this is due to LDM being used for the RM reconstruction task for the first time. In subsequent research, the relevant pre-trained weights of VAE can be directly used without re-training, which will significantly streamline future research efforts.}
Moreover, techniques such as NN compression and efficient inference methods like denoising diffusion implicit models (DDIM) \cite{song2020denoising} can be leveraged to notably enhance efficiency, raising a trade-off between performance and efficiency, which stands as a promising direction for future research.

Furthermore, it is noteworthy that almost all existing NN-based RM construction methods, including this paper, concentrate on predicting pathloss from a BS to a specific location, known as one-to-any (O2X) RM. However, there is also another type of RM that is any-to-any (X2X) RM scenario, which aims at obtaining pathloss between any two points through their positions. X2X RM construction poses a challenge with sampling-based approaches, as these methods typically require a fixed BS location to estimate pathloss of any other position to the fixed BS. In contrast, by adopting a sampling-free RM construction approach in this paper and incorporating the location of BS as input for the diffusion model as part of the prompt, it is feasible to predict pathloss between any two points by adjusting the prompt of the BS position, thereby enabling the construction of X2X RM.

\begin{table}[t]
\centering
\caption{Inference Time and Memory Consuming. Results in bold red and underlined blue highlight the highest and second highest, respectively.}
\vspace{-2pt}
\captionsetup{font={small}, skip=8pt}
\resizebox{0.5\linewidth}{!}{
\begin{tabular}{@{}c|c|c@{}}
\toprule
\textbf{Method} & \textbf{Average time (s)} & \textbf{Memory consuming (MB)} \\ \midrule
RME-GAN & 0.042 & 2923 \\
UVM-Net & {\color[HTML]{00009B} \underline{0.095}} & {\color[HTML]{9A0000} \textbf{8738}} \\
RadioUNet & 0.056 & 3927 \\
RadioDiff (Ours) & {\color[HTML]{9A0000} \textbf{0.6}} & {\color[HTML]{00009B} \underline{6062}} \\ \bottomrule
\end{tabular}
}
\label{tab3}
\vspace{-12pt}
\end{table}

\section{Conclusion}\label{sec-7}
In this paper, we have proposed RadioDiff, a diffusion-based RM generative model to effectively construct the RM. By incorporating various techniques, including AFTs and the decoupled diffusion model, RadioDiff can construct accurate and sharp RM effectively. Extensive experiments demonstrate the qualitative and quantitative superiority of the proposed RadioDiff. As the first application of diffusion models to RM construction tasks, RadioDiff sets a new benchmark for future technological advancements. In future work, we will focus on how to leverage the diffusion model to generate the environment features based on the sparse RM information.

\bibliography{ref}
\bibliographystyle{IEEEtran}

\ifCLASSOPTIONcaptionsoff
  \newpage
\fi

\end{document}